\documentclass[10pt,twocolumn]{article}

\usepackage[margin=0.72in,columnsep=0.24in]{geometry}
\usepackage[T1]{fontenc}
\usepackage[utf8]{inputenc}
\usepackage{lmodern}
\usepackage{microtype}
\usepackage{parskip}
\usepackage{enumitem}

\usepackage{amsmath,amssymb}
\usepackage{booktabs}
\usepackage{tabularx}
\usepackage{array}
\usepackage{makecell}
\usepackage{multirow}
\usepackage{graphicx}
\usepackage{xcolor}
\usepackage{tikz}
\usetikzlibrary{arrows.meta,positioning,fit,calc}
\usepackage{pgfplots}
\pgfplotsset{compat=1.18}
\usepackage{siunitx}
\usepackage{listings}

\usepackage[hidelinks]{hyperref}
\usepackage[nameinlink,noabbrev]{cleveref}
\usepackage{url}
\setlist[itemize]{leftmargin=*,topsep=2pt,itemsep=1pt,parsep=0pt}
\setlist[enumerate]{leftmargin=*,topsep=2pt,itemsep=1pt,parsep=0pt}

\definecolor{eqblue}{HTML}{2457A7}
\definecolor{eqlightblue}{HTML}{EAF1FB}
\definecolor{eqgreen}{HTML}{2F7D5C}
\definecolor{eqlightgreen}{HTML}{EAF6F0}
\definecolor{eqgold}{HTML}{B8860B}
\definecolor{eqlightgold}{HTML}{FBF4DD}
\definecolor{eqgray}{HTML}{5B6573}
\definecolor{eqlightgray}{HTML}{F3F5F7}

\lstdefinestyle{svsnippet}{
    basicstyle=\ttfamily\scriptsize,
    columns=fullflexible,
    keepspaces=true,
    frame=single,
    rulecolor=\color{eqlightgray},
    backgroundcolor=\color{eqlightgray!55},
    xleftmargin=2pt,
    xrightmargin=2pt,
    aboveskip=4pt,
    belowskip=4pt,
    showstringspaces=false
}

\newcommand{\dataset}{\textsc{EquivSVA}}

\newcommand{\repo}{\url{https://github.com/aditigupta96/EquivSVA}}
\newcommand{\release}{\url{https://github.com/aditigupta96/EquivSVA/releases/tag/v2.0}}

\title{\vspace{-1.2em}\textbf{EquivSVA: A Formally Verified Dataset of Behavioral Assertions Across Equivalent RTL Implementations}}
\author{\textbf{FNU Aditi}}
\date{}

\begin{document}
\maketitle
\vspace{-1.4em}

\begin{abstract}
Large language models are increasingly used to generate SystemVerilog Assertions from natural-language specifications and register-transfer-level designs. Existing datasets and benchmarks support important goals such as large-scale training, formal evaluation, specification-to-assertion generation, and mutation-based testing. A complementary need is to study whether a generated assertion captures externally observable behavior or depends on incidental details of one RTL implementation. We present \dataset, a formally verified dataset organized around \emph{behavior families}. Each family contains four structurally distinct RTL implementations of the same externally observable behavior, shared interface-level gold properties, three controlled mutants, and formal-validation evidence. \dataset\ contains 120 behavior families across 12 categories, 480 reference RTL implementations, 914 gold properties, and 360 mutants. Every final family passes a fixed 17-job validation suite covering RTL equivalence, gold-property proofs, property reachability, mutant distinguishability, and gold-property checks on mutants. We also provide fixed family-safe train, development, and test splits. As a small demonstration of the analyses enabled by the dataset, we evaluate the publicly released, Apache-2.0-licensed Qwen2.5-Coder-7B-Instruct model on the held-out test split. Of 293 interface-only generated properties, 93 are formally sound, and the number of sound properties varies across equivalent implementations for 14 of 24 test families. These results illustrate how behavior-family organization can support controlled studies of assertion-generation robustness without requiring changes in intended functionality. The dataset, generators, validation scripts, and case-study artifacts are publicly released at \repo.
\end{abstract}

\section{Introduction}
\label{sec:introduction}

Assertion-based verification provides a concise way to express design intent and to check temporal and safety properties of digital hardware. In practice, writing useful SystemVerilog Assertions (SVAs) requires both an understanding of the intended behavior and careful attention to clocking, reset semantics, signal relationships, and temporal structure. This has motivated substantial work on automatic assertion generation, including rule-based and learned translation from natural-language specifications \cite{aditi2022hybrid,aditi2023validatable}, specification-driven LLM frameworks \cite{fang2024assertllm}, benchmark suites for formal-verification tasks \cite{kang2024fveval,pulavarthi2025assertionbench}, and datasets for training assertion-generation models \cite{menon2025vert,wu2026codevsva}.

A recurring challenge is that RTL admits many implementations of the same behavior. State encodings, conditional structure, helper expressions, combinational factorization, and update style can all change while the external behavior remains unchanged. A model that generates a correct assertion for one implementation may therefore behave differently when presented with another semantically equivalent implementation. Prior work has examined this issue through semantics-preserving RTL transformations \cite{aditi2026robustness}. That study motivates a reusable dataset in which behavioral equivalence is not an auxiliary perturbation applied after dataset construction, but a first-class organizing principle.

We introduce \dataset, a dataset whose basic unit is a \emph{behavior family}. Each family begins from a machine-readable behavioral specification and contains four structurally distinct reference RTL implementations intended to realize the same externally observable behavior. The family also includes shared interface-level gold properties, three controlled behavior-changing mutants, and formal evidence used to validate the final artifacts. This organization makes it possible to ask whether an assertion-generation method responds consistently to different implementations of the same behavior while holding the intended semantics fixed.

The goal of \dataset\ is complementary to existing resources. VERT emphasizes large-scale RTL--SVA training data \cite{menon2025vert}; FVEval organizes formal-verification tasks at multiple levels of abstraction \cite{kang2024fveval}; AssertionBench provides curated designs with formally verified assertions for model evaluation \cite{pulavarthi2025assertionbench}; CodeV-SVA develops RTL-grounded data synthesis for specialized NL-to-SVA models \cite{wu2026codevsva}; Veri2 organizes formally filtered RTL--SVA pairs for fine-tuning \cite{goradia2026veri2}; and AssertLLM2 provides real-world designs, structured specifications, golden RTL, and systematically mutated buggy RTL for realistic assertion-generation evaluation \cite{wu2026assertllm2}. \dataset\ adds a behavior-family representation that places multiple formally equivalent RTL implementations, shared gold properties, and controlled mutants under one dataset unit.

The main contributions are:
\begin{itemize}
    \item A public dataset of 120 behavior families across 12 hardware-control categories, containing 480 reference RTL implementations, 914 gold properties, and 360 controlled mutants.
    \item A family-centered representation in which four structurally distinct RTL implementations are formally checked for equivalent externally observable behavior.
    \item A fixed validation protocol that checks RTL equivalence, gold-property correctness, property reachability, mutant distinguishability, and whether each mutant violates at least one property in the family gold-property harness.
    \item Family-safe, category-stratified train, development, and test splits designed to prevent implementation variants from the same behavior appearing across different splits.
    \item A held-out Qwen2.5-Coder-7B-Instruct case study demonstrating syntax, formal soundness, mutation sensitivity, and variation across equivalent RTL implementations.
\end{itemize}

The released artifact is available at \repo, with the v2.0 snapshot at \release.

\section{Related Work}
\label{sec:related}

\subsection{Assertion generation from specifications}

Earlier work explored translating natural-language requirements into SVAs using hybrid rule-based and machine-learning methods \cite{aditi2022hybrid}. A later approach introduced a validation loop that translated generated SVAs back into natural-language statements and then regenerated assertions to assess consistency with the original specification \cite{aditi2023validatable}. AssertLLM subsequently studied complete specification documents and used multiple LLM stages for structure extraction, signal mapping, and assertion generation \cite{fang2024assertllm}. These works establish assertion generation as a structured reasoning task spanning natural-language intent, hardware signals, and temporal logic.

More recent systems have expanded both the generation methodology and the evaluation setting. SANGAM uses multi-stage specification processing and Monte Carlo tree self-refinement for SVA generation \cite{gupta2025sangam}; Spec2Assertion targets pre-RTL assertion generation with progressive regularization \cite{wu2025spec2assertion}; and AssertGen connects specification-level verification objectives to RTL signals before producing assertions \cite{lyu2025assertgen}. These approaches focus on improving the generation process itself, whereas \dataset\ focuses on a reusable data representation for controlled evaluation across implementation variants.

\subsection{Datasets and benchmarks}

VERT constructs augmented RTL--SVA training examples from open-source HDL and is designed to support fine-tuning of open-source language models \cite{menon2025vert}. FVEval provides three formal-verification subtasks, spanning natural-language-to-SVA generation and assertion generation directly from RTL, together with tool-backed evaluation \cite{kang2024fveval}. AssertionBench contains 100 curated Verilog designs from OpenCores and formally verified assertions generated using GoldMine and HARM, enabling quantitative comparison of LLMs for assertion generation \cite{pulavarthi2025assertionbench}. CodeV-SVA uses RTL-grounded bidirectional data synthesis to create training data for specialized assertion-generation models \cite{wu2026codevsva}.

Veri2 is a formally filtered RTL--SVA dataset that organizes generated pairs into quality tiers and reports 2,954 modules and 18,494 assertions in its verified tier \cite{goradia2026veri2}. AssertLLM2 provides 83 real-world designs across 13 categories together with structured specifications, golden RTL, systematically mutated buggy RTL, and evaluation spanning syntax, formal provability, coverage, and mutation-based bug detection \cite{wu2026assertllm2}. These resources address complementary questions about training data, realistic specifications, tool-backed evaluation, and bug detection.

\begin{table*}[t]
\centering
\caption{Representative related datasets and benchmarks, described by how their examples are organized and what they are designed to support. The table emphasizes complementary design goals rather than ranking the resources.}
\label{tab:related}
\small
\begin{tabularx}{\textwidth}{@{}p{0.12\textwidth}p{0.23\textwidth}X X@{}}
\toprule
\textbf{Work} & \textbf{Primary purpose} & \textbf{How examples are organized} & \textbf{Tool-backed / formal role} \\
\midrule
VERT \cite{menon2025vert} & Training data for RTL-to-SVA generation & Augmented open-source RTL--SVA training pairs & Evaluates generated assertions and downstream fine-tuned models \\
FVEval \cite{kang2024fveval} & Benchmarking LLM capabilities in hardware formal verification & Task instances for NL2SVA-Human, NL2SVA-Machine, and Design2SVA & Jasper-based evaluation framework \\
AssertionBench \cite{pulavarthi2025assertionbench} & Quantitative comparison of LLM assertion generation & 100 curated Verilog designs with formally verified assertion sets & Formal verification is used to establish benchmark assertions \\
CodeV-SVA \cite{wu2026codevsva} & Training specialized NL-to-SVA models & RTL-grounded synthetic NL--SVA training pairs produced by bidirectional synthesis & Uses semantic filtering and benchmark evaluation \\
Veri2 \cite{goradia2026veri2} & Formally filtered RTL--SVA training data & RTL--SVA pairs organized into All, Syntax Pass, and Verified tiers & JasperGold filtering of syntax and formal validity \\
AssertLLM2 \cite{wu2026assertllm2} & Realistic assertion-generation benchmark & Real-world designs with specifications, golden RTL, and systematically mutated buggy RTL & Syntax, proof, coverage, and mutation-based evaluation \\
\textbf{EquivSVA} & Behavior-centered dataset for implementation-robust assertion research & \textbf{Behavior families with 4 equivalent RTLs, shared gold properties, and 3 controlled mutants} & \textbf{Dataset-wide equivalence, proof, cover, and mutant checks} \\
\bottomrule
\end{tabularx}
\end{table*}

A closely related motivation is robustness under semantics-preserving RTL rewriting. Prior work demonstrated that LLM-generated assertions can change in quality when the same behavior is represented by transformed RTL \cite{aditi2026robustness}. \dataset\ turns that observation into a dataset-level abstraction: multiple equivalent implementations are stored directly in each family and can be reused across models, prompts, and evaluation protocols.

\section{Dataset Design}
\label{sec:design}

\subsection{Behavior families}

The core design decision in \dataset\ is to organize data around behavior rather than around isolated RTL files. A family is represented conceptually as
\begin{equation}
\mathcal{F} = \left(B, \{R_i\}_{i=1}^{4}, P, M, E\right),
\end{equation}
where $B$ is a machine-readable behavior specification, $R_i$ are four reference RTL implementations, $P$ is the shared set of gold behavioral properties, $M$ is a set of three controlled mutants, and $E$ is formal-validation evidence.

The four reference implementations are intentionally different in structure. Depending on the family type, variation includes state encoding, case versus nested conditional control, factorized flag logic, sequential versus ternary updates, and function-based update expressions. The intended external behavior is held constant. Gold properties are defined over module-interface signals rather than implementation-specific internal state so that the same behavioral specification can be applied across variants.

\begin{figure*}[t]
\centering
\begin{tikzpicture}[
    node distance=7mm and 6mm,
    every node/.style={font=\small},
    box/.style={rounded corners=3pt, draw=eqblue, very thick, fill=eqlightblue, align=center, minimum height=1.25cm, text width=2.55cm},
    greenbox/.style={rounded corners=3pt, draw=eqgreen, very thick, fill=eqlightgreen, align=center, minimum height=1.25cm, text width=2.55cm},
    goldbox/.style={rounded corners=3pt, draw=eqgold, very thick, fill=eqlightgold, align=center, minimum height=1.25cm, text width=2.55cm},
    graybox/.style={rounded corners=3pt, draw=eqgray, very thick, fill=eqlightgray, align=center, minimum height=1.25cm, text width=2.55cm},
    arrow/.style={-{Stealth[length=2.5mm]}, very thick, draw=eqgray}
]
\node[box] (spec) {\textbf{Behavior specification}\\machine-readable intent};
\node[greenbox, right=of spec] (rtl) {\textbf{RTL family}\\4 structurally distinct reference implementations};
\node[goldbox, right=of rtl] (props) {\textbf{Gold properties}\\shared, interface-level behavioral assertions};
\node[graybox, right=of props] (mutants) {\textbf{Controlled mutants}\\3 behavior-changing variants};
\node[box, right=of mutants] (formal) {\textbf{Formal validation}\\equivalence, proof, cover, and mutant checks};

\draw[arrow] (spec) -- (rtl);
\draw[arrow] (rtl) -- (props);
\draw[arrow] (props) -- (mutants);
\draw[arrow] (mutants) -- (formal);

\node[draw=eqgreen, dashed, rounded corners=3pt, fit=(rtl)(props)(mutants), inner sep=5pt, label={[font=\scriptsize\bfseries,text=eqgreen]below:Behavior family artifact}] {};

\node[below=11mm of props, font=\small\bfseries, text=eqgray] (outcome) {120 families $\cdot$ 480 reference RTLs $\cdot$ 914 gold properties $\cdot$ 360 mutants};
\draw[-{Stealth[length=2mm]}, thick, draw=eqgray] (formal.south) |- (outcome.east);
\end{tikzpicture}
\caption{\dataset\ construction and validation pipeline. The family abstraction keeps intended behavior fixed while exposing multiple implementation structures and controlled behavior-changing mutants.}
\label{fig:pipeline}
\end{figure*}
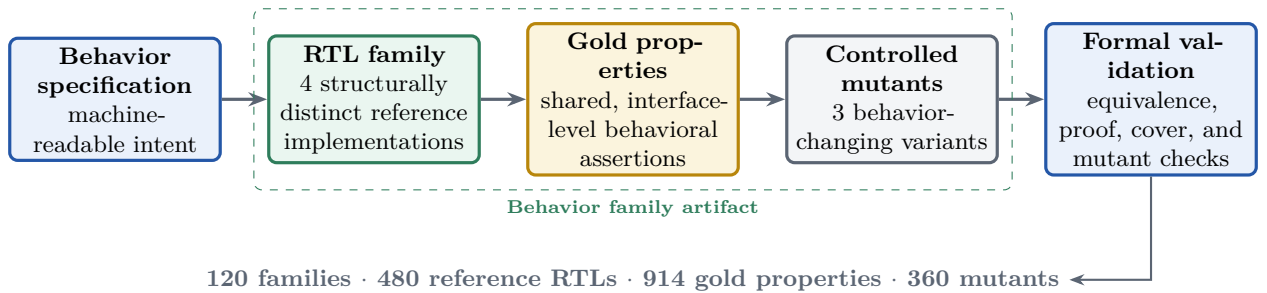

\subsection{Behavior categories and model types}

The dataset contains 12 categories chosen to cover recurring control and small-state behaviors: arbiter, counter, FIFO control, handshake, interrupt control, mode controller, protocol controller, pulse/event, rate limiter, saturating arithmetic, sequence detector, and timer/watchdog. Each category contains exactly 10 families. The final dataset includes 80 finite-state-machine families, 27 register-rule families, and 13 multi-register-rule families.

This category balance is deliberate. It makes category-level comparisons straightforward and prevents the overall metrics from being dominated by one frequently generated behavior class. It should not be interpreted as an estimate of how often these structures appear in industrial RTL.

\subsection{Gold properties}

The dataset contains 914 gold behavioral properties. Of these, 131 are invariants and 783 are next-cycle implications. Family-level property counts range from 5 to 13, with a mean of 7.62. Properties are written to describe externally observable functionality and avoid implementation-specific internal state names. This constraint is central to the family abstraction: a property should continue to represent the intended behavior even when the internal implementation changes.

\subsection{Concrete family example}
\label{sec:family-example}

As a concrete example, \texttt{timer\_0007} is a three-bit countdown timer. A \texttt{load} input sets the observable \texttt{remaining} value to seven; a \texttt{tick} decrements a nonzero count; otherwise the count holds. The derived output \texttt{expired} is asserted exactly when \texttt{remaining} is zero. The family contains four reference implementations---canonical, sequential, ternary, and function-oriented update styles---generated from the same rule-level specification. Its three controlled mutants respectively ignore \texttt{load}, ignore \texttt{tick}, and decrement without requiring \texttt{tick}. Thus, the family changes implementation structure while preserving one intended interface behavior, and changes behavior only in the explicitly labeled mutant artifacts.

Two representative gold properties for this family illustrate the property forms used in the corpus. The first is a next-cycle implication associated with the load rule; the second is an invariant relating the derived output to the observable count:

\begin{lstlisting}[style=svsnippet]
property p_load;
  @(posedge clk) disable iff (rst)
    load |=> (remaining == 3'd7);
endproperty

a_load: assert property (p_load);

property p_expired;
  @(posedge clk) disable iff (rst)
    expired == (remaining == 3'd0);
endproperty

a_expired: assert property (p_expired);
\end{lstlisting}

These examples are interface-level: neither property depends on a particular state encoding, helper signal, or internal register name that differs among the four reference implementations. Appendix~\ref{app:examples} gives one representative behavior family from each of the 12 dataset categories.

\begin{table*}[t]
\centering
\caption{Composition of \dataset\ across the 12 behavior categories. Every family contains four reference RTL implementations and three controlled mutants.}
\label{tab:composition}
\small
\begin{tabular}{@{}lrrrrr@{}}
\toprule
\textbf{Category} & \textbf{Families} & \textbf{Reference RTLs} & \textbf{Gold properties} & \textbf{Mutants} & \textbf{Props./family} \\
\midrule
Arbiter                 & 10 & 40 & 103 & 30 & 10.30 \\
Counter                 & 10 & 40 & 58  & 30 & 5.80 \\
FIFO control            & 10 & 40 & 63  & 30 & 6.30 \\
Handshake               & 10 & 40 & 74  & 30 & 7.40 \\
Interrupt control       & 10 & 40 & 71  & 30 & 7.10 \\
Mode controller         & 10 & 40 & 91  & 30 & 9.10 \\
Protocol controller     & 10 & 40 & 78  & 30 & 7.80 \\
Pulse/event             & 10 & 40 & 70  & 30 & 7.00 \\
Rate limiter            & 10 & 40 & 62  & 30 & 6.20 \\
Saturating arithmetic   & 10 & 40 & 64  & 30 & 6.40 \\
Sequence detector       & 10 & 40 & 109 & 30 & 10.90 \\
Timer/watchdog          & 10 & 40 & 71  & 30 & 7.10 \\
\midrule
\textbf{Total}          & \textbf{120} & \textbf{480} & \textbf{914} & \textbf{360} & \textbf{7.62} \\
\bottomrule
\end{tabular}
\end{table*}

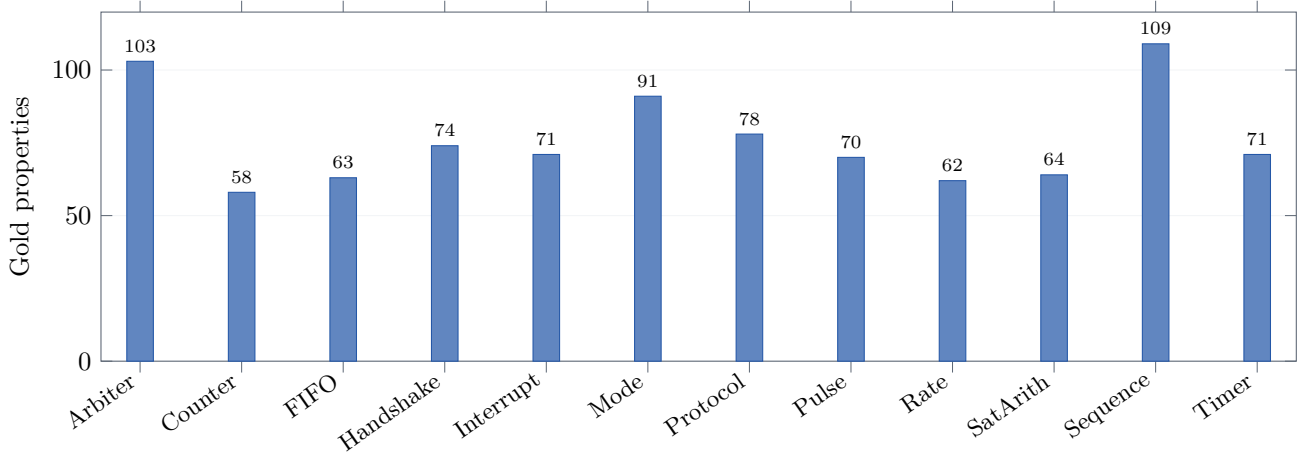
\begin{figure*}[t]
\centering
\begin{tikzpicture}
\begin{axis}[
    width=0.97\textwidth,
    height=6.2cm,
    ybar,
    bar width=10pt,
    ymin=0,
    ylabel={Gold properties},
    symbolic x coords={Arbiter,Counter,FIFO,Handshake,Interrupt,Mode,Protocol,Pulse,Rate,SatArith,Sequence,Timer},
    xtick=data,
    x tick label style={rotate=35,anchor=east,font=\small},
    ymajorgrids=true,
    grid style={draw=eqlightgray},
    axis line style={draw=eqgray},
    tick style={draw=eqgray},
    nodes near coords,
    nodes near coords style={font=\scriptsize},
    enlarge x limits=0.035,
]
\addplot[fill=eqblue!72,draw=eqblue] coordinates {
    (Arbiter,103) (Counter,58) (FIFO,63) (Handshake,74)
    (Interrupt,71) (Mode,91) (Protocol,78) (Pulse,70)
    (Rate,62) (SatArith,64) (Sequence,109) (Timer,71)
};
\end{axis}
\end{tikzpicture}
\caption{Gold-property counts by behavior category. Category family counts are fixed at 10, so the variation reflects the number of behavioral properties associated with each family set rather than category size.}
\label{fig:properties-by-category}
\end{figure*}

\section{Dataset Construction}
\label{sec:construction}

\subsection{Specification-driven generation}

Families are generated from explicit machine-readable specifications rather than by independently sampling unrelated RTL files. Three generator paths are used: finite-state machines, single-register rule systems, and multi-register rule systems. The generator emits multiple implementation styles from the same behavior description. Because variants share the same intended semantics but differ in control structure and coding form, they can be used to study implementation sensitivity without changing the target behavior.

For FSM families, implementation styles include canonical case statements, one-hot state encodings, nested conditionals, and factored flag logic. Register-rule and multi-register-rule families use canonical, sequential, ternary, and function-oriented update forms. The precise coding differences vary by family so that the dataset does not reduce to a single text-rewrite pattern.

\subsection{Controlled mutants}

Each family contains three mutants designed to change behavior in a controlled way. Mutation operators depend on the family structure and include dropped transitions, forced exits, ignored control conditions, missing updates, altered clear or enable behavior, incorrect saturation, and similar localized semantic changes. Mutants are not intended to model the full distribution of industrial hardware bugs. Their purpose is to provide known behavior-changing alternatives against which assertions can be tested.

\subsection{Diversity audit}

The final corpus contains no exact normalized behavioral clones according to the released diversity audit. The audit also identifies a small number of parameter- or shape-similar groups, which are retained because they remain distinct behavior families. Template metadata is diverse across categories, with one repeated FIFO metadata template. We therefore describe the corpus as 120 \emph{behavior families}, rather than claiming 120 unique behavioral archetypes.

\section{Formal Validation}
\label{sec:formal}

Formal validation is used as a quality-control layer for the final dataset. The released flow uses Yosys and SymbiYosys-based infrastructure \cite{wolf2013yosys,symbiyosys}, with SMT backends including Bitwuzla \cite{niemetz2023bitwuzla}. Multi-register equivalence checks use ABC/PDR where that flow is more reliable for the generated design class.

Every final family passes a fixed 17-job validation suite. Three jobs compare alternate reference implementations against the canonical implementation. Four proof jobs exercise the family gold-property set across the four reference implementations, and four cover jobs check the associated reachability witnesses. Three jobs establish that each controlled mutant is distinguishable from the reference behavior. The final three jobs run the family gold-property harness against each mutant in bounded model-checking mode and require an expected assertion failure. A mutant therefore passes this check only when at least one gold property produces a counterexample.

A validation \emph{job} can contain multiple assertions. In particular, each of the four gold-property proof jobs instantiates one reference implementation together with the complete property set for that family. The 480 proof jobs in \cref{tab:formal} therefore collectively prove every one of the 914 gold properties on each of the four reference implementations in its family; the job count should not be interpreted as the number of individual properties. The same distinction applies to cover jobs, which contain the family reachability witnesses.

\begin{table}[t]
\centering
\caption{Formal-validation jobs associated with the final dataset. Every final family passed its complete 17-job suite.}
\label{tab:formal}
\footnotesize
\setlength{\tabcolsep}{3pt}
\begin{tabular}{@{}p{0.60\columnwidth}rr@{}}
\toprule
\textbf{Validation check} & \textbf{Jobs/family} & \textbf{Pass} \\
\midrule
RTL equivalence                  & 3 & 360/360 \\
Gold-property proofs             & 4 & 480/480 \\
Property reachability / cover    & 4 & 480/480 \\
Mutant distinguishability        & 3 & 360/360 \\
Gold-property checks on mutants  & 3 & 360/360 \\
\midrule
\textbf{Total}                   & \textbf{17} & \textbf{2040/2040} \\
\bottomrule
\end{tabular}
\end{table}

The 2,040 total in \cref{tab:formal} summarizes the final per-family validation records. During dataset development, failed or malformed families were repaired and then revalidated. Accordingly, we use the precise statement that \emph{every final family passed its 17-job validation suite}, rather than implying that the entire corpus was accepted in a single uninterrupted first-pass run.

The formal results are relative to the encoded synchronous clock and reset semantics. Validation traces begin with the required reset condition, after which inputs are unconstrained according to the family harness. Equivalence therefore means equality of the defined externally observable outputs under the shared harness assumptions.

\section{Splits and Release}
\label{sec:splits}

The v2.0 release provides fixed family-safe, category-stratified train, development, and test splits. All four RTL variants of a family remain in the same split. This prevents a model from seeing one implementation of a behavior during training and another implementation of the same behavior during evaluation.

\begin{table}[t]
\centering
\caption{Fixed \dataset\ v2.0 splits. Every category contributes 6 train, 2 development, and 2 test families.}
\label{tab:splits}
\small
\begin{tabular}{@{}lrrrr@{}}
\toprule
\textbf{Split} & \textbf{Families} & \textbf{RTLs} & \textbf{Properties} & \textbf{Mutants} \\
\midrule
Train & 72 & 288 & 544 & 216 \\
Dev   & 24 & 96  & 178 & 72 \\
Test  & 24 & 96  & 192 & 72 \\
\bottomrule
\end{tabular}
\end{table}

The split was generated deterministically. Expansion families are used for development and test, while legacy pilot families are assigned to training where prior prompt or baseline work could have exposed them. This choice is conservative with respect to possible experiment leakage from earlier prototype work.

The public release contains the dataset manifest, split file, generators, construction and validation scripts, task-export code, baseline inference code, syntax/formal/mutant evaluators, and the v2 case-study result files. The repository is licensed separately for code and dataset artifacts, with Apache-2.0 for source code and CC BY 4.0 for the dataset.

\section{Case Study: Qwen2.5-Coder-7B}
\label{sec:case-study}

We include a small case study to demonstrate how the family structure can be used in model evaluation. The purpose is not to provide a comprehensive model ranking. We evaluate Qwen2.5-Coder-7B-Instruct \cite{hui2024qwen25coder}, whose upstream release is distributed under the Apache License 2.0, on the untouched test split: 24 families and 96 RTL inputs. Inference uses the public 4-bit MLX checkpoint \nolinkurl{mlx-community/Qwen2.5-Coder-7B-Instruct-4bit}; the exact checkpoint identifier is also stored in the released run metadata. Decoding is greedy. The prompt requests interface-only behavioral SVAs and prohibits implementation-specific internal signals.

Generated outputs are evaluated in stages. First, syntax is checked. Next, properties that reference only interface signals and fall within the supported lowering subset are translated into formal monitors. The run produced 371 extracted assertions; 357 were supported by the lowering pipeline, and 293 were both lowerable and interface-only. A property is counted as formally sound when the proof succeeds on its source RTL. Sound properties are then checked against the three family mutants. Finally, results are aggregated by behavior family to measure variation across the four equivalent implementations.

For transparency, strict raw prompt-format compliance was 0/96 because outputs did not exactly obey the requested bare-declaration format. The evaluator therefore separates formatting adherence from syntactic validity of the extracted SVA content. After the evaluator's normalization/extraction step, 68/96 tasks produced syntactically valid SVA.

Let $P_{\mathrm{int}}$ be the set of interface-only generated properties and $P_{\mathrm{pass}}$ the subset formally proven on their source RTL. We report property soundness as
\begin{equation}
\mathrm{Soundness} = \frac{|P_{\mathrm{pass}}|}{|P_{\mathrm{int}}|}.
\end{equation}

For family-level implementation sensitivity, let $s_i$ denote the number of sound properties produced from RTL variant $i$. A family is counted as variant-sensitive when the four values are not all equal:
\begin{equation}
\mathrm{Sensitive}(\mathcal{F}) = \mathbf{1}\!\left[|\{s_1,s_2,s_3,s_4\}| > 1\right].
\end{equation}

\begin{table}[t]
\centering
\caption{Held-out Qwen2.5-Coder-7B-Instruct case study on 24 test families (96 RTL inputs).}
\label{tab:qwen}
\small
\begin{tabular}{@{}lr@{}}
\toprule
\textbf{Metric} & \textbf{Result} \\
\midrule
Syntax-valid tasks & 68/96 (70.8\%) \\
Interface-only properties & 293 \\
Formally sound properties & 93/293 (31.7\%) \\
Tasks with $\geq$1 sound property & 45/96 (46.9\%) \\
Families sound on all 4 RTLs & 8/24 (33.3\%) \\
Variant-sensitive families & 14/24 (58.3\%) \\
Detected property--mutant pairs & 16/279 (5.7\%) \\
Unique controlled mutants detected & 11/72 (15.3\%) \\
\bottomrule
\end{tabular}
\end{table}

Of the 293 interface-only properties, 93 (31.7\%) are formally sound. Forty-five of 96 RTL tasks produce at least one sound property. At the family level, eight of 24 families produce at least one sound property for all four equivalent implementations, while eight families produce none for any implementation. The remaining eight produce sound properties for only a subset of implementations.

A notable family-level observation is that the number of sound properties changes across equivalent implementations for 14 of 24 families (58.3\%). This does not by itself identify the cause of the difference, nor does it imply that one coding style is globally more difficult. It demonstrates the type of controlled analysis enabled by storing multiple equivalent implementations under the same family label.

Mutation testing provides an additional view of usefulness. Across 279 checks pairing a sound generated property with a family mutant, 16 checks detect the behavioral change. These detections cover 11 of the 72 unique mutants in the test families. The detected mutants span multiple categories, including arbiter, FIFO control, handshake, pulse/event, sequence detector, and timer/watchdog. We treat this as a demonstration metric rather than a complete measure of assertion quality.

\section{Discussion}
\label{sec:discussion}

\subsection{What the family abstraction enables}

A conventional RTL-to-SVA example asks whether a property is correct for one implementation. A behavior family supports additional questions while keeping the intended semantics fixed. For example, researchers can measure whether a model produces sound properties for all variants, whether the number or type of properties changes by implementation style, whether generated assertions transfer across family members, and whether mutation sensitivity is stable across implementations. The same family structure can also support training objectives that encourage representation invariance or contrastive reasoning across equivalent designs.

\subsection{Complementarity with prior resources}

The contribution of \dataset\ is not that other datasets should be reorganized in the same way. Different resources serve different needs. Large training corpora are useful for fine-tuning; real-world specification benchmarks are useful for evaluating practical generation settings; mutation-based resources test bug-detection behavior; and tool-backed benchmarks provide rigorous correctness signals. \dataset\ contributes a controlled equivalence dimension that can be used alongside these existing directions. This is why \cref{tab:related} describes each work by its organization and purpose rather than by a checklist of missing features.

\section{Public Artifacts and Data Provenance}
\label{sec:provenance}

The EquivSVA behavior specifications, generated RTL, assertions, mutants, scripts, validation artifacts, and evaluation outputs used in this work are publicly released. The model case study uses only Qwen2.5-Coder-7B-Instruct, whose upstream release is distributed under the Apache License 2.0. The study does not use proprietary models, non-public datasets, internal source code or infrastructure, customer data, or confidential information. The released repository contains the artifacts needed to reproduce the reported dataset statistics and case-study evaluation.

\section{Limitations}
\label{sec:limitations}

\dataset\ is intentionally controlled and therefore has several limitations. First, the families are programmatically generated rather than mined directly from industrial code bases. This provides precise semantics and repeatable formal validation, but it may not capture the full structural complexity, naming conventions, long-range dependencies, or specification ambiguity present in production RTL.

Second, the 12 categories emphasize control logic, small state machines, counters, handshakes, timers, and related behaviors. The dataset does not attempt comprehensive coverage of large datapaths, caches, coherent interconnects, deeply pipelined arithmetic units, or full protocol stacks. Third, the gold-property distribution is dominated by next-cycle implications and invariants. Longer-horizon liveness and richer temporal sequences remain an important direction for future extensions.

Fourth, the equivalent variants are generated from common machine-readable specifications and generator families. Formal equivalence establishes behavioral agreement under the harness assumptions, but generated variants can still share stylistic regularities not representative of independently authored RTL. Fifth, controlled mutants are designed to be behavior-changing and formally distinguishable; they should not be interpreted as a statistically representative sample of hardware defects.

Finally, the Qwen2.5-Coder-7B experiment is a single-model case study. Its purpose is to demonstrate dataset usage and family-level metrics, not to establish a model leaderboard. Broader multi-model comparisons, prompt strategies, fine-tuning experiments, and deeper mutation analyses are left for subsequent work.

\section{Conclusion}
\label{sec:conclusion}

We presented \dataset, a formally verified dataset organized around behavioral equivalence families. The v2.0 release contains 120 families across 12 categories, 480 reference RTL implementations, 914 gold behavioral properties, and 360 controlled mutants. Each family pairs four structurally distinct but formally equivalent implementations with shared interface-level properties and behavior-changing mutants. Every final family passes a 17-job validation suite, and fixed family-safe splits support reproducible training and evaluation.

A held-out Qwen2.5-Coder-7B-Instruct case study illustrates the central use case: assertion quality can vary even when the intended behavior is unchanged. By making equivalent implementations a first-class dataset element, \dataset\ provides a reusable basis for studying whether assertion-generation systems capture behavioral intent rather than implementation-specific structure. The dataset and accompanying tools are publicly available at \repo.

\appendix
\section{Representative Families by Category}
\label{app:examples}

The following examples give one representative behavior family from each dataset category. They are illustrative rather than canonical definitions; the released dataset contains ten families per category.

\begin{itemize}
    \item \textbf{Arbiter} (\texttt{arbiter\_0006}, FSM): sticky two-client arbitration in which an active grant is held until release.
    \item \textbf{Counter} (\texttt{counter\_0006}, register rules): modulo-six counter with synchronous clear and enable control.
    \item \textbf{FIFO control} (\texttt{fifo\_0006}, multi-register rules): elastic one-entry queue supporting same-cycle pop-and-replace behavior.
    \item \textbf{Handshake} (\texttt{handshake\_0006}, FSM): request remains active until acknowledgment, followed by request release.
    \item \textbf{Interrupt control} (\texttt{interrupt\_0006}, FSM): masked interrupt request is latched when enabled and cleared by acknowledgment.
    \item \textbf{Mode controller} (\texttt{mode\_controller\_0006}, FSM): locked, ready, and active operating modes with lock/unlock control.
    \item \textbf{Protocol controller} (\texttt{protocol\_controller\_0002}, FSM): command/response protocol that waits for a response before reporting completion.
    \item \textbf{Pulse/event} (\texttt{pulse\_event\_0002}, FSM): one-shot event pulse with rearming after the triggering level is released.
    \item \textbf{Rate limiter} (\texttt{rate\_limiter\_0002}, register rules): token state with refill priority and bounded token capacity.
    \item \textbf{Saturating arithmetic} (\texttt{saturating\_arithmetic\_0002}, register rules): incrementing value that saturates at five and supports synchronous clear.
    \item \textbf{Sequence detector} (\texttt{sequence\_detector\_0002}, FSM): overlapping serial detector for the bit pattern \texttt{110}.
    \item \textbf{Timer/watchdog} (\texttt{timer\_0007}, register rules): countdown timer loaded to seven and decremented on tick until expiration.
\end{itemize}

\section*{Artifact Availability}
The public dataset, generators, validation scripts, evaluation code, and case-study artifacts are available at \repo. The frozen v2.0 snapshot used for this paper is available at \release.

\section*{AI Use Disclosure}
Generative AI tools were used to assist with software development and manuscript preparation. All technical content and results were reviewed and validated by the author.

\bibliographystyle{plain}
\bibliography{references}

\end{document}